\documentclass[lettersize,journal,twoside]{IEEEtran}
\usepackage{amsmath,amsfonts}
\usepackage{algorithmic}
\usepackage{algorithm}

\usepackage{xcolor}
\usepackage{array}
\usepackage[caption=false,font=normalsize,labelfont=sf,textfont=sf]{subfig}
\usepackage{textcomp}
\usepackage{stfloats}
\usepackage{url}
\usepackage{verbatim}
\usepackage{caption}
\usepackage{graphicx}
\usepackage{cite}
\usepackage{multirow}
\usepackage{booktabs}
\usepackage{xcolor}
\usepackage[x11names,table]{xcolor}
\usepackage{colortbl}
\usepackage{makecell}
\usepackage{tabularray}
\usepackage{pifont}
\usepackage{gensymb}
\usepackage{etoolbox}
\definecolor{gr}{HTML}{E3EDD9}
\definecolor{bg}{HTML}{F0F0F0}
\renewcommand{\tablename}{Table}

\makeatletter
\patchcmd{\@makecaption}{\scshape}{}{}{}
\makeatother

\begin{document}

\markboth{IEEE ROBOTICS AND AUTOMATION LETTERS. PREPRINT VERSION. ACCEPTED MAY, 2026}
{Shen \MakeLowercase{\textit{et al.}}: SurfSurg6D: Geometry Consistent Dense Correspondence for Textureless Surgical Instrument Pose Estimation}

\title{SurfSurg6D: Geometry Consistent Dense Correspondence\\
for Textureless Surgical Instrument Pose Estimation}

\author{
Daiyun Shen$^{1}$,
Shuojue Yang$^{1}$,
Chang Han Low$^{1}$,
Qian Li$^{1}$,
Mengya Xu$^{2}$,
Qi Dou$^{2}$,~\IEEEmembership{Senior Member,~IEEE,}
and Yueming Jin$^{1}$,~\IEEEmembership{Member,~IEEE}
\thanks{Manuscript received: March 25, 2026; Accepted: May 10, 2026. This paper was recommended for publication by Editor-in-Chief Tamim Asfour and Editor Jessica Burgner-Kahrs upon evaluation of the Associate Editor and reviewers' comments. This work was supported by Ministry of Education Tier 2 grant, Singapore (T2EP20224-0028), and Ministry of Education Tier 1 grant, Singapore (23-0651-P0001). (Corresponding author: Yueming Jin)}%
\thanks{$^{1}$D. Shen, S. Yang, C. H. Low, Q. Li and Y. Jin are with the National University of Singapore, Singapore
(\texttt{ymjin@nus.edu.sg}).}%
\thanks{$^{2}$M. Xu and Q. Dou are with the Chinese University of Hong Kong, HKSAR, China.}%
}

\maketitle

\begin{abstract}
 Surgical instrument pose estimation provides crucial information for promising applications, including autonomous robotic surgery, skill assessment, and standardization of surgical workflow. However, this task remains highly challenging due to high precision requirements, frequent occlusions, textureless instruments, scarcity of depth information and very limited annotated data. These constraints often lead to unsatisfactory performance when employing general object pose estimation approaches to surgical scenarios. To address these issues, we first construct a new dataset \textbf{SynSurg6D}, to alleviate the data shortage in this task. We further propose \textbf{SurfSurg6D}, a dense-correspondence framework tailored for surgical instrument pose estimation. Experimental results on the \textbf{SurgRIPE}, \textbf{EndoVis2018} and \textbf{SurgPose} datasets demonstrate that the introduction of our generated dataset \textbf{SynSurg6D} is able to diversify the pose distributions, thus enhancing the performance of existing approaches. Furthermore, \textbf{SurfSurg6D} outperforms existing methods, providing a robust solution for precise and efficient RGB-only pose estimation. The code and dataset will be open-sourced at https://github.com/StackingDataYeti/SurfSurg6D.

\end{abstract}

\begin{IEEEkeywords}
Pose estimation, Surgical scene synthesis, Surgical robotics, Contrastive learning, Data generation
\end{IEEEkeywords}

\section{Introduction}
\IEEEPARstart{S}{\MakeLowercase{urgical}} instrument pose estimation plays a critical role in a wide range of emerging applications in robotic minimally invasive surgery (RMIS). By accurately determining the rotation and translation of surgical tools in real time, it provides essential spatial information to support the standardization of surgical procedures and enhance downstream tasks such as skill assessment\cite{rarp}, error detection   \cite{jigsaw}
 , and surgical workflow analysis   \cite{transsv,reviewforswa,svrcnet}
. Furthermore, precise pose estimation helps mitigate errors in forward kinematics data, thereby enabling more accurate robot motion control and facilitating semi-automation of surgical subtasks.

Despite its potential, surgical instrument pose estimation faces several challenges. Hardware solutions such as depth cameras, kinematic sensors, or electronic trackers are often impractical in operating rooms \cite{nasir}. Data-driven approaches can bypass these constraints \cite{yip}, but require large annotated datasets, which remain scarce in surgery. Due to the cable-driven mechanism of RMIS robots \cite{cable1,cable2}, pose labels derived from forward kinematics are unreliable \cite{davinci,raven}. Previous works \cite{SurgRIPE,handeye} calibrate poses using key-dot panels but restrict instrument motion and pose diversity. Simulation-based datasets \cite{datageneration} increase scale but still suffer from a domain gap with real surgical images.

\begin{figure}[!t]
\centering
\includegraphics[width=3in]{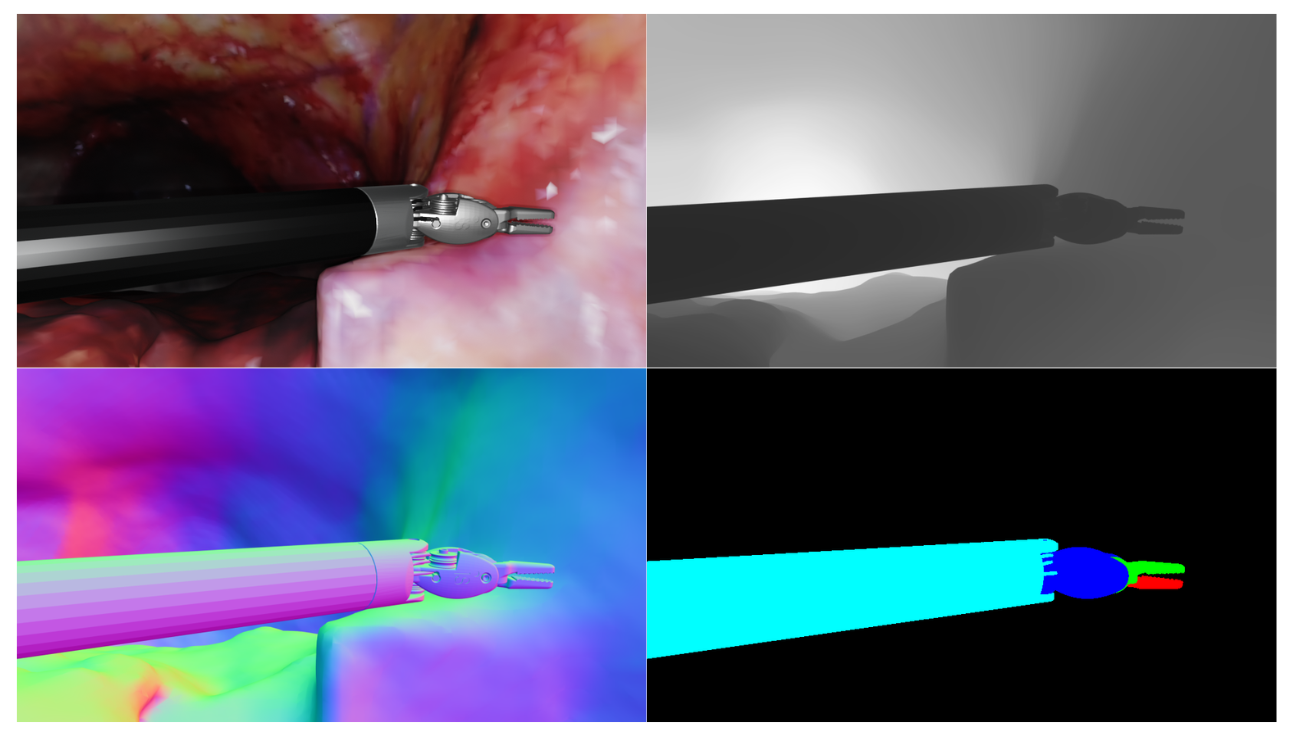}
\caption{Examples of generated synthetic dataset SynSurg6D. Modalities including RGB-D images, segmentation maps, normal maps, and 6-DoF poses of each instrument part are provided. }
\vspace{-5mm}
\label{fig_introduction}
\end{figure}

   Existing general pose estimation  approaches can be categorized into three streams: regression-based, template-based and correspondence-based methods. Regression-based methods such as GDR-net \cite{GDR-net} simplify the pipeline by predicting the 6D pose directly from RGB or RGB-D inputs. MRC-Net \cite{mrcnet} 
utilizes a coarse-to-fine paradigm that first predicts a rough pose, then renders the image and concatenate it with the origin one to regress the final pose. However ,   the training data shortage in the surgical domain weakens the model's robustness. 

Template-based methods such as MegaPose \cite{megapose}, FoundPose \cite{foundpose}, and FoundationPose \cite{foundationpose} recently achieve remarkable progress in 6-DoF unseen object pose estimation in the natural domain. Leveraging large and diverse synthetic datasets and pretrained vision foundation models like DINOv2 \cite{dinov2}, these methods perform well with textured objects under uniform lighting. However, their effectiveness remains limited in surgery. This underperformance can be attributed to insufficient training data in surgical domain and limitations of template-matching mechanisms. Template-based methods rely on matching real images to pre-rendered templates\cite{focalpose}, highly sensitive to visual feature discrepancies between rendered templates and real imaging conditions. In surgical environments, frequent occlusion and dynamic lighting exacerbate this gap, leading to degraded performance on textureless instruments.

Correspondence-based methods are better suited for surgical instrument pose estimation, owing to their ability to establish 2D-3D correspondence relationships even under challenging conditions such as occlusion and limited texture.    Some previous works of surgical instrument pose estimation\cite{artnet,kp_yip,lu2022unified} 
rely on sparse geometric features, including keypoints and object boundary lines. They offer high inference speed but are somehow sensitive to occlusion. Increasing keypoint numbers may alleviate the difficulty, yet requiring a lot of manual selection and annotations. Other works\cite{daniel,instrumentsplatting}
utilize part-level masks and features to perform more robust matching and tracking, but these features are not fine-grained enough to ensure rotational precision, and sometimes manual initialization is needed. Dense correspondence methods like Surfemb \cite{surfemb}, offer robust and fine-grained 2D-3D surface-level associations across the entire object, making them better suited for handling geometric information and self-occluding objects commonly encountered in surgical environments.    Their success  depends on the dataset scale to ensure pose distribution diversity, as well as the full exposure of instrument surface, so that dense 2D-3D correspondence can be established on the whole object. And these methods lack the attention for the similarity of fragments on the textureless object's surface, which may cause confusion and lead to significant errors in difficult cases.  

In this work, we construct a rendering pipeline to generate \textbf{SynSurg6D}, a large-scale synthetic surgical instrument pose dataset comprising over 60K rendered images across 6 commonly used surgical instruments with ground truth pose annotations. Incorporating diverse backgrounds and realistic lighting variations, the dataset narrows the gap between synthetic renderings and real surgical images, making it a practical resource for data-driven methods aiming for application in reality.  Examples of the generated dataset modalities are shown in Fig. \ref{fig_introduction}.  We then introduce \textbf{SurfSurg6D}, a dense correspondence-based framework tailored for textureless surgical instrument pose estimation. To enhance the discrimination of similar surface fragments, we design a distance-based penalization strategy that promotes hard negative pair mining during contrastive learning, improving the model’s ability to distinguish locally similar but geometrically different points. We also propose a consistency loss    that guarantees  the learned implicit embeddings    to respect the underlying 3D geometry, discouraging local folding  of the surface    and improving embedding space distribution. Experimental results on the dataset SurgRIPE \cite{SurgRIPE} show that SurfSurg6D achieves reliable performance under significant occlusion, while maintaining    competitive  speed necessary for practical RMIS applications. In summary, our contributions are as follows:

\textbullet{} We propose \textbf{SynSurg6D}, a large-scale synthetic dataset of surgical instruments. It supplies abundant images with accurate 6-DoF pose annotations,    multiple-type data variations and authentic rendering quality.

\textbullet{} We introduce \textbf{SurfSurg6D}, a correspondence-based framework that integrated two mechanisms: geometric distance-based hard negative mining strategy to enhance correspondence learning, and embedding consistency regularization to better model textureless surfaces.

\textbullet{} We conduct extensive evaluation on    three  surgical benchmark datasets. Experimental results show that our method achieves superior performance than other state-of-the-art methods, and the introduction of \textbf{SynSurg6D} brings consistent improvement to all methods. 

\begin{table*}[!ht]
  \centering
    \caption{Dataset comparison. SurgRIPE obtains 6D pose annotations by calibrating an attached key-dot panel (which is inpainted in the final image). SurgPose\cite{surgpose} provides 2D keypoint annotations by attaching fluorescent markers, but limited keypoints are provided, which makes it hard to derive 6D pose annotations. Barragan \emph{et~al.}\cite{datageneration} provides the synthetic images and 6D poses of a needle, but its image quality is limited. EndoVis 2018 provides images in real surgical scenarios, but no pose annotation is provided.  }
  \resizebox{0.6\linewidth}{!}{%
    \begin{tabular}{lccccc}
      \toprule[1pt]
        Dataset    & Size & Instrument & Modality & 6D Pose annotation &Real \\

      \midrule[0.5pt]
      SurgRIPE~\cite{SurgRIPE}      &    2K   & 2  & RGB & $\checkmark$ & $\checkmark$ \\
      SurgPose\cite{surgpose}       &   60K & 6 & RGB-D & $\times$ & $\checkmark$\\
      Barragan \emph{et~al.}~\cite{datageneration} & 8K & 2 &RGB-D & $\checkmark$  & $\times$ \\
      EndoVis2018~\cite{EndoVis2018}        & 6K  & 6 & RGB-D & $\times$ & $\checkmark$ \\
      SynSurg6D       & 60K & 6 & RGB-D & $\checkmark$ & $\times$ \\
      \bottomrule[1pt]
    \end{tabular}%
  }
  \vspace{0.25mm}

  \label{tab:dataset_compare}
 \end{table*}

\section{Methods}

\begin{figure*}[!ht]
    \centering
    \hspace{-3mm}\includegraphics[scale=0.40
    ]{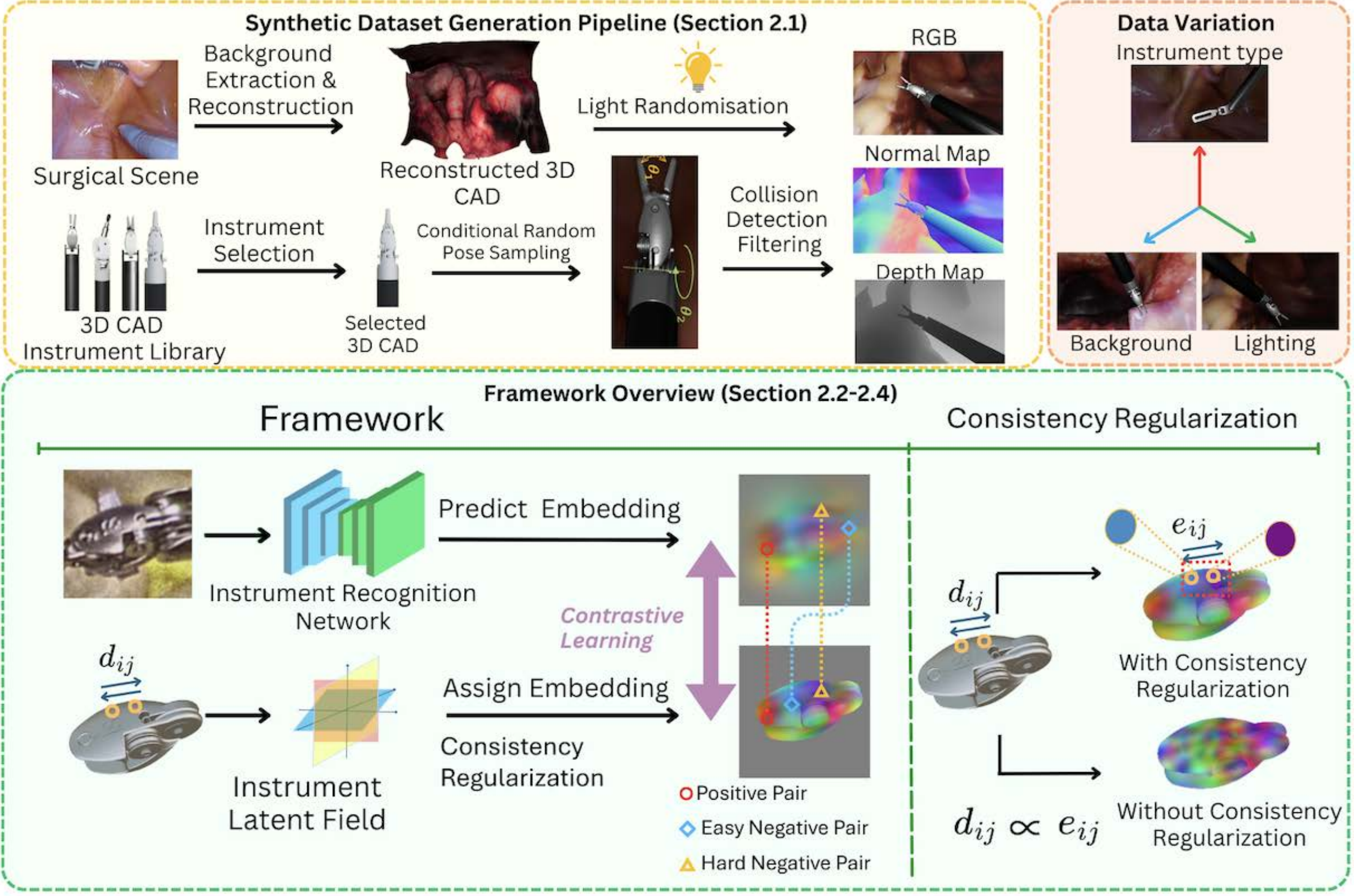}
    \caption{  Overview of the synthetic dataset generation pipeline and the SurfSurg6D framework. (a) Scene reconstruction and dataset generation. (b) The framework consists of an Instrument Latent Field that encodes instrument surfaces, and an Image Recognition Network that predicts per-pixel embeddings. (c) We further introduce distance-aware hard negative mining and a geometric-embedding consistency loss (Sec. II-C–II-D) 
    }
    \label{fig:mainfig}
\end{figure*}

\subsection{New Dataset Generation}

Prior work \cite{foundationpose} shows that training data diversity and scale are vital for pose estimation performance. Due to the cable-driven nature of RMIS instruments, precise ground-truth 6-DoF poses are hard to acquire in real scenes. Synthetic datasets would be an alternative to relieve the data shortage.  In the general pose estimation domain, works including Benchmark of Pose estimation\cite{bop} and \cite{2012textureless} provide automatic approaches of rendering multiple objects with different poses. In the surgical domain, \cite{datageneration} provides an automated data generation pipeline for 6D pose estimation of surgical instruments. However, the unsatisfactory rendering quality limits its potential for transferring to real images.  

Surgical scene reconstruction methods help fill the blank in a realistic surgical background. EndoSurf \cite{endosurf} provides the approach of reconstructing accurate surfaces of endoscopic environments of EndoNerf \cite{endonerf} datasets. The reconstructed mesh model would function as the realistic background for data generation. Meanwhile, BlenderProc \cite{blenderproc} provides a realistic rendering pipeline with editable lighting and scene settings, as well as diverse modalities including segmentation, depth, normals, and ground truth poses

\begin{figure}[!t]
    \centering
    \includegraphics[width=3in]{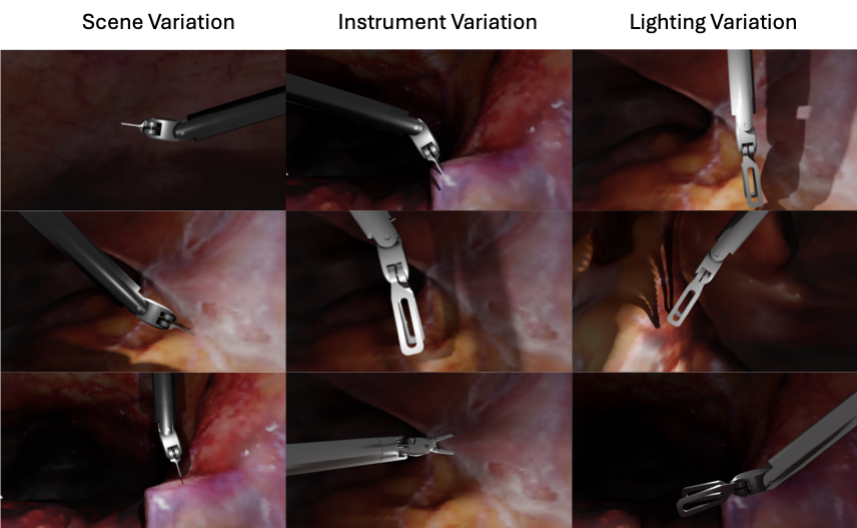}
    \caption{Examples of data variation, including background scene, instrument type and lighting. }
    \vspace{-6mm}

    \label{fig_variation}
    
    \end{figure}

We generate the synthetic dataset by reconstructing 3D anatomical models, carefully positioning the instruments to reflect realistic surgical configurations, and applying randomized lighting and background variations to cover a wide range of intraoperative scenarios, see Fig. \ref{fig_variation}.  Specifically, we apply a remote-center-of-motion (RCM) constraint, requiring the instrument shaft to pass through a fixed trocar point in each reconstructed scene. The translational degrees of freedom are limited such that the tool tip remains within a 30–100 mm depth range from the local tissue surface, and the angles of the distal wrist and clips are sampled and limited based on the articulated structure derived from the kinematic specifications of clinical robotic systems. In addition, we reject samples where the instrument leaves the field-of-view or where the visible part of the wrist falls below a certain threshold.  Our instrument library consists of 6 instruments: Large Needle Driver, Scissors, Small Clip Applier, Cadiere Forceps, Biopsy Forceps, and Permanent Cautery Spatula. The Large Needle Driver CAD model is derived from the opensource dVRK project\cite{dvrk}
, and the others are obtained from public GrabCAD repositories \cite{grabcad}.  
The pipeline consists of three stages:
\begin{itemize}
  \item  Background extraction, reconstruction, and instrument selection
  \item  Conditional pose sampling and collision detection
  \item  Light randomization and rendering
\end{itemize}
   We reconstruct 6 surgical scenes 3D mesh models based on the EndoNeRF\cite{endonerf} dataset with the method of EndoSurf\cite{endosurf} in advance. For the first stage, we randomly select one instrument and one scene model. In the second stage, we apply a minor jitter to the scene models and set the selected instrument's pose according to the scene model's geometry, ensuring the instrument and the scene model are in the central part of the view.    For each articulated instrument, we explicitly model the distal wrist as a low-DOF kinematic chain. In this work we focus on the last four DOFs starting from the shaft under an RCM constraint: insertion along the shaft, axial roll, and two wrist bending angles. During sampling, we first determine the trocar position, then sample the insertion depth and roll angle, and finally jitter the two wrist angles within anatomically plausible ranges. We limit the roll angle with the range of ±30 degrees and the two wrist angles with the range of 0-60 degrees. This approximates the clinically observed motion of the end-effector while keeping the kinematics tractable. 
Collision detection is performed in the whole scene to exclude mesh penetration circumstances and limited exposure of instruments.  We perform mesh-level collision filtering to ensure physically plausible configurations. For each sampled pose, we exclude both instrument–tissue collisions and self-collisions between distinct instrument components to avoid generating unrealistic samples. We implement these checks using Blender’s BVH-accelerated mesh intersection queries as exposed by BlenderProc \cite{blenderproc}. Any configuration exhibiting mesh penetration is discarded. In addition, we enforce a visibility constraint by projecting the instrument mask to the image plane and rejecting samples whose visible projected area is below 15\% of the full (unoccluded) silhouette, preventing heavily occluded or ambiguous training examples.

    During the final stage, randomization of lighting conditions will be performed per frame.    The  light location and direction will be altered along with the trocar point and shaft direction, and the light intensity distribution will be shifted. The dataset is arranged in the format of Benchmark for 6D Object Pose Estimation \cite{bop}.
    
\vspace{-4mm}
\subsection{Framework Overview}

Our goal is to estimate the 6-DoF pose (3 DoF rotation + 3 DoF translation) of a    rigid part of  instrument based on an image crop \(I\in\mathbb{R}^{H\times W\times3}\) according to the provided bounding boxes. We achieve this by learning the correspondence between the pixel $u$ in the image and the point $c$ on the surface of the instrument. The framework mainly consists of two networks to learn such correspondence: one instrument latent field $M_{\phi}$ that implicitly assigns embeddings to normalized coordinates on the instrument surface, and one image recognition network $F_{\theta}$ that recognizes each pixel's embedding of the given image:

\begin{itemize}
  \item \emph{instrument latent field} \(\mathcal{M}_\phi: \mathbb{R}^3\to\mathbb{R}^E\), mapping 3D normalized surface points \(c_i\in\mathcal{V}\subset\mathbb{R}^3\) to key embeddings
  \begin{equation} \label{eq:1}
      E_i = \mathcal{M}_\phi(c_i)\in\mathbb{R}^E\;,\;\forall\,c_i\in{\mathcal{V}}.
 \end{equation}

  \item \emph{image recognition network} \(\mathcal{F}_\theta: \mathbb{R}^{H\times W\times3}\to\mathbb{R}^{H\times W\times E}\), producing a 2D embedding map
  \begin{equation} \label{eq:2}
\quad
      \widetilde{E_j} = F_\theta[u_j]\in\mathbb{R}^E\;,\;\forall\,u_j\in\mathcal{U},
  \end{equation}

\end{itemize}
where $\mathcal{V}$ denotes the set of surface points of the instrument, and $\mathcal{U}$ denotes the set of pixels in the image. Note that both $E_i$ and $\widetilde{E_i}$ are normalized embeddings.
\noindent

In the training stage, we construct positive pairs (for example, coordinate $c_i$ and corresponding pixel $u_i$) and negative pairs ($c_i$ and all other irrelevant pixels) to apply contrastive learning techniques, making embeddings of positive pair $\widetilde{E_i}$ and $E_i$ are close in the latent space, and vice versa. Such a goal can be achieved by applying InfoNCE loss \cite{infonce} 
 over sampled pixels \(\tilde{\mathcal{U}}\subset\mathcal{U}\):
\begin{equation} \label{eq:4}
  \mathcal{L}_{\mathrm{InfoNCE}}
  = -\frac{1}{|\tilde{\mathcal{U}}|}\sum_{u_j\in\tilde{\mathcal{U}}}
    \log
    \frac{\exp\!\bigl(\langle \widetilde{E_i}, E_{i}\rangle\bigr)}{\sum_{c_j\in\tilde{\mathcal{V}}\cup\{c_{i}\}}
                \exp\!\bigl(\langle \widetilde{E_i}, E_j\rangle\bigr)}
\end{equation}

Here, we note that each negative pair is treated equally. However, some hard negative pairs have not been adequately emphasized in the original framework \cite{surfemb}
. We explore this in the next section.

The network can further benefit from the ground truth mask by adding an additional prediction dimension of Image Recognition Network to predict $\hat{p}_{u}$, the probability the pixel $u$ is in the mask $M$, supervised with the binary cross-entropy loss:
\begin{equation} \label{eq:5}
L_{M}=
-\frac{1}{\lvert \mathcal{U}\rvert}
\sum_{u \in \mathcal{U}}
\Bigl[
y_{u}\,\log\bigl(\hat{p}_{u}\bigr)
\;+\;
\bigl(1 - y_{u}\bigr)\,\log\bigl(1 - \hat{p}_{u}\bigr)
\Bigr]
\end{equation}
\noindent
where $y_{u}$ denotes if pixel {u} is in the object mask.

The final loss is 
\begin{equation}
L=L_{\mathrm{InfoNCE}}+L_{M}    
\end{equation}

This objective corresponds to the original Surfemb formulation\cite{surfemb} 
and will be referred to as our baseline loss.

 At the inference stage, by matching $E_i$ and $\widetilde{E_i}$, we construct the correspondence pair $\{u_i, c_i\}$ according to a combination of mask score and correspondence score. Then we may apply RANSAC-PnP \cite{pnp1,pnp2} 
 to solve the final pose.

The paradigm mainly follows Surfemb\cite{surfemb}, and the method has been shown to be effective, especially for textureless objects. However, as previously noted, the role of hard negative pairs has been underemphasized. Also, the instrument latent field lacks regularization for distribution, leading to large pose errors even when slight correspondence errors occur. To enhance the stability of instrument pose estimation and better address issues of surface similarity-induced confusion for textureless instruments, we introduce a hard-negative mining strategy and latent field consistency loss to improve the existing framework, which provide the model with higher robustness for textureless objects and prove to be simple and effective. Fig. \ref{fig:mainfig} shows the overall pipeline of our work. 
\vspace{-4mm}
\subsection{Hard Negative Pair Mining for Pixel-point Pair}

Hard negative mining is introduced by \cite{hardnegative}, with the intuition that not all negative pairs are equally informative, and those hard negative pairs should be prioritized to improve the contrastive learning process. In the context of pixel-point correspondence, a negative pair consists of a query embedding $\widetilde{E_i}$ related to the pixel $u_i$ ($u_i$ corresponds to the surface point $c_i$) and a non-matching surface point embedding $E_j$ related to the point $c_j$. 

Some negative pairs, where the pixel and the surface point are far apart in 3D ($c_i$ and $c_j$ are far away from each other), tend to be trivial to discriminate, contributing relatively little to the fine-grained correspondence learning of the network. In contrast, spatially nearby non-matching pixel-and-point pairs are more likely to induce confusion, under the intuition that visually these pairs are relatively similar, no matter of color or lighting. These negative pairs are harder for the network, and emphasizing them encourages sharper   and more localized features    (see examples in Fig.\ref{fig:mainfig}).  We implement this via a distance-aware penalty that up-weights hard negatives:

\begin{equation}
{p_{ij}(\alpha) = (1-\alpha\log(||c_j-c_i||+1))} 
\end{equation}

and the penalized InfoNCE loss may be presented as:

{\small
\begin{equation}
L_{\mathrm{penNCE}}(\alpha)
= -\log
\frac{%
  \exp\!\bigl(\langle \widetilde E_i,\,E_i\rangle\bigr)
}{%
  \exp\!\bigl(\langle \widetilde E_i,\,E_i\rangle\bigr)
  + \sum_{j=1}^N p_{ij}(\alpha) \,\exp\!\bigl(\langle \widetilde E_i,\,E_j\rangle\bigr)
}\
\end{equation}
}

Note that $c_i$ and $c_j$ are normalized coordinates, and we may guarantee that $p_{ij}$ is positive when choosing a reasonable $\alpha$. 
By penalizing the contribution of easy negatives in the softmax denominator, the network is forced to focus its representation capacity on distinguishing the most ambiguous, spatially proximate false matches. This strategy accelerates convergence and yields more precise pixel–point alignments, especially in scenes with cluttered backgrounds or low‐texture regions. 
\vspace{-4mm}
\subsection{Localized Correspondence Consistency for Implicit Embedding Model}
In Surfemb, the instrument latent field $M_\phi$ is optimized along with $F_\theta$ solely by the correspondence-matching objective without additional regularization. While this direct training can succeed in learning useful pixel–point associations, it also allows the embedding manifold to become fragmented in localized regions, meaning that two surface points that are close in 3D space might end up far apart in the latent space, or vice versa.
During inference, this means that even if the predicted embedding $\widetilde{E_i}$
for pixel ${u_i}$
does not exactly match the true embedding $E_i$	of its corresponding surface point $c_i$,
 it can still be close to some other embedding $E_j$
  of a different surface point $c_j$. If $c_j$ lies far from $c_i$ in the 3D space, choosing $c_j$
as the top-scoring surface point can introduce a large error when solving for the pose via PnP.

 To prevent this, we 
 impose a geometric-embedding consistency constraint, requiring the embedding distance of two points should correlate with the geometric distance, \textit{i.e.}, if two surface points $c_i$ and $c_j$ are geometrically close, then their embeddings $E_{i}$ and $E_{j}$ should also be close    (see in Fig.\ref{fig:mainfig}).  We enforce such requirements by adding the consistency loss, which can be formulated in the form of:

\begin{equation}
    L_{con}= \frac{2}{N(N-1)} \sum_{i=1}^N \sum_{j=i+1}^N w_{ij} (||c_i-c_j|| - ||E_i-E_j||)^2
\end{equation}
where 
\begin{equation}
w_{ij}= max(m,\exp(-\lambda||c_i-c_j||)  )
\end{equation}
$\lambda$ denotes the distance scale factor, $m$ denotes min weight constant to prevent the weight from vanishing, and $N$ denotes the number of surface points in \(\tilde{\mathcal{V}}\). We  use an exponential decay with a floor $m$ so that nearby points receive higher consistency weights while distant points are not completely ignored. This choice is able to stabilize training better than a hard distance threshold or a linear decay. We  choose $\lambda$=10 and $m$=0.01 in our experiments after trial and error. 

This consistency term encourages a smooth and continuous embedding field, so that even when the predicted embedding $\widetilde{E_i}$
  is slightly off, it will still rank nearby surface points—thereby reducing the chance of picking a geometrically distant $c_j$,
  and improving overall pose accuracy.

Finally, we integrate the regularization into our full training objective:
\begin{equation}
    L(\alpha,\beta)=L_{\mathrm{penNCE}} (\alpha)  + L_M + \beta L_{con}
\end{equation}

Compared to Eq.5, the original InfoNCE term is replaced by the penalized one in Eq.7,  where $\alpha$ controls the strength of hard negative re-weighting and $\beta$ balances the consistency regularization.

\begin{table*}[!ht]

    \centering
     \caption{Overall performance comparison on the SurgRIPE dataset consisting of occlusion scenes where occlusion frequently occurs and non-occlusion scenes.  We include ICL and IGTUM (SurgRIPE challenge methods) as additional baselines.  For  MRC-Net,  Surfemb and GDR-Net, we train them from scratch from publicly available code. For foundation models like MegaPose and FoundPose, we do zero-shot inference from publicly available checkpoints.   }       \LARGE

  \resizebox{\linewidth}{!}{
        \begin{tabular}{cc||cccc||cccc||cccc}
        \toprule[1.5pt]
Training     & \multirow{2}{*}{Method}& \multicolumn{4}{c||}{SurgRIPE Non-occ} & \multicolumn{4}{c||}{SurgRIPE occ} & \multicolumn{4}{c}{Avg}  \\
        \cmidrule{3-14}

         setting &  & {\large RE(\degree) $\downarrow$} &{\large TE(mm)$\downarrow$} & {\large ADD-10 (\%) $\uparrow$}& {\large Avg ACC(0-5 mm)(\%)$\uparrow$}& {\large RE(\degree) $\downarrow$} & {\large TE(mm) $\downarrow$} & {\large ADD-10 (\%) $\uparrow$}&{\large Avg ACC(0-5mm)(\%)$\uparrow$} & {\large RE(\degree) $\downarrow$} & {\large TE(mm) $\downarrow$} & {\large ADD-10 (\%) $\uparrow$} &{\large Avg ACC (0-5mm)(\%) $\uparrow$}\\
        \midrule[1pt]
         zero-shot &MegaPose\cite{megapose} & 132.30 & 12.40 & 0.00 & 0.13 & 125.96 & 12.00 & 0.00& 0.14 & 129.83  & 12.24 & 0.00 & 0.13 \\
           &FoundPose\cite{foundpose}&112.34 & 28.86 & 2.20 &5.49 &116.79 & 25.37 & 2.05  &5.43 & 114.07 & 27.50 & 2.14 & 5.46 \\
        \midrule[1pt]        real&ICL\cite{SurgRIPE}&   57.17 & 63.32 & 18.23& 26.57& 28.45  & 66.52 & 20.59& 28.99& 45.98 & 64.56 & 19.15 &27.51\\ &IGTUM\cite{SurgRIPE}&   5.18 & \textbf{2.56} & \textbf{41.82} & \textbf{56.69} & 10.71  & 5.40 & 36.55&50.44 & 7.33 & 3.66 & 39.76 & 54.25\\&MRC-Net\cite{mrcnet}&   14.84  & 3.42 &37.00 &49.85 &20.99  & 4.71 & 21.85& 34.81&17.24  &3.92  &31.09  &43.99\\&GDR-Net\cite{GDR-net}  & 11.29 & 3.65 & 28.95& 39.76 & 15.03 & 4.13 & 28.99&48.95 & 12.75 & 3.84 & 28.97& 43.33  \\
       &Surfemb\cite{surfemb}&   5.21 & 2.91 & 39.67 & 54.48 & 10.36 & 4.12 & 44.53 & 57.53 & 7.22 & 3.38 & 41.56 & 55.68 \\    
       &\textbf{SurfSurg6D} & \textbf{5.05} & 2.94 & 41.55 & 54.74 & \textbf{8.57} & \textbf{3.61} & \textbf{44.96} & \textbf{57.87} &\textbf{6.42} & \textbf{3.20} & \textbf{42.89}& \textbf{55.95}   \\
      \midrule[1pt]
      syn+real &MRC-Net& 11.17   & 3.14 & 43.16&55.14 &  13.10 & 3.66 & 28.57& 43.32& 11.92 & 3.34 & 37.47 &50.53\\&GDR-Net & 8.01 & 3.23 & 32.43 &47.01  & 11.78 & 4.21 & 31.51 &42.66 & 9.48 & 3.61 & 32.07 & 45.31\\  
        &Surfemb  & 4.78 & \textbf{2.28} & 44.77 & 59.40& 10.44 & 3.33 & 45.38 & 58.32 & 6.98 & 2.69 & 45.01 & 58.97\\
        &\textbf{SurfSurg6D } & \textbf{4.78} & 2.44 & \textbf{45.04} & \textbf{60.06} &
\textbf{9.42} & \textbf{2.92} & \textbf{49.15} & \textbf{60.22} &
\textbf{6.59} & \textbf{2.63} & \textbf{47.46} & \textbf{60.12}\\
        \bottomrule[1.5pt]        \end{tabular}
    }
       \vspace{0.25mm}
 \label{tab:main-results}
\end{table*}

 \section{Experiments and Results} \subsection{Datasets and Evaluation Metrics}
We employ    three  datasets for evaluation: SurgRIPE\cite{SurgRIPE}   ,  EndoVis2018\cite{EndoVis2018}  and SurgPose\cite{surgpose}
. SurgRIPE is a public surgical instrument pose estimation dataset, capturing surgical instruments with tissue backgrounds in real scenes with pose annotations, camera intrinsics, and CAD models of specific parts. We utilize this dataset to evaluate the effectiveness of the method and the efficacy of the generated dataset brought to the model with respect to the metric in pose estimation. EndoVis2018 is a dataset that contains real surgical scenes with ground truth masks of specific parts of the instruments.    SurgPose is a dataset that precisely annotates the keypoints on the instruments. We utilize the two datasets  to assess the similarity of prediction and ground truth in 2D projection  on mask and keypoint level. 

For pose estimation,  following the SurgRIPE benchmark\cite{SurgRIPE}.  we choose common metrics of evaluating the accuracy of pose estimation: rotation error(RE), translation error(TE),   ADD-10 accuracy (a sample is considered correct when ADD is lower than 10\% diameter of the object)    and average accuracy in 0-5mm (Avg acc) as final  metrics.

 For 2D projection  in EndoVis2018 , we employ the Dice score \cite{dice} 
between ground truth segmentation map and prediction projection map  to assess the projection loss  on mask level. Our goal is to assess each approach's performance in real clinical images. However, the Dice score is influenced by occlusions and thus only loosely reflects 3D pose quality.

 So we employ SurgPose to further identify each approach's performance with more quantifiable metrics. For the wrist part of Large Needle Driver, SurgPose annotates 3 keypoints (No.2,3,6). We annotate the corresponding keypoints on the 3D mesh and project them to 2D plane with predicted instrument pose and known camera intrinsics. We follow the evaluation conventions in SurgPose\cite{surgpose}
, choosing the metric mean average precision (mAP) over multiple OKS thresholds(0.5:0.05:0.95), where OKS stands for Object Keypoint Similarity, defined by COCO\cite{coco}
. The object scale in OKS is defined by the bounding box of the wrist part, and we set $\sigma$ as 0.1 for all three points. 
\vspace{-4mm}
\subsection{Implementation Details}
The instrument latent field employs SIREN MLP\cite{sirenmlp}, and the image recognition network employs ResNet18\cite{resnet} pretrained on ImageNet with the U-Net architecture. The optimizer is Adam\cite{whoisadam} with a learning rate of $3e-4$ for the image recognition network and $3e-5$ for the instrument latent field, which is consistent with the original Surfemb setting\cite{surfemb}. We provide ground truth bounding boxes of SurgRIPE, EndoVis2018, and SynSurg6D for the network.

For Surfemb and SurfSurg6D, we use a batch size of 20 and total iteration steps of 13000. For the loss coefficient in SurfSurg6D, we take $\alpha$=1 and $\beta$=1.  We choose 1024 positive query-key pairs and 1024 negative points. For synthetic-real-blended training, we choose to balance the dataset ratio to $1:1$ in each batch by performing data augmentation , as a higher syn-real ratio may result in domain shift and bring performance degradation.  We train and run our models on one NVIDIA RTX Ada 5000 GPU.
\vspace{-4mm}
\subsection{Comparison with State-of-the-Art Methods}
\begin{table}[t]
\centering
\caption{2D projection evaluation on EndoVis2018 and SurgPose.}
\resizebox{0.9\linewidth}{!}{
\begin{tabular}{lccc}
\toprule
Setting & Method & Dice (EndoVis2018)$\uparrow$ & mAP (SurgPose)$\uparrow$\\
\midrule
\multirow{2}{*}{zero-shot}
& MegaPose & 0.5463 & 0.0030 \\
& FoundPose & 0.7110 & 0.0976 \\
\midrule
\multirow{4}{*}{real}
& MRC-Net & 0.7168 & 0.2678 \\
& GDR-Net & 0.7322 & 0.4693 \\
& Surfemb & 0.7566 & 0.5797 \\
& \textbf{SurfSurg6D} & 0.7555 & \textbf{0.6756} \\
\midrule
\multirow{4}{*}{syn+real}
& MRC-Net & 0.7550 & 0.6006 \\
& GDR-Net & 0.7653 & 0.5201 \\
& Surfemb & 0.7753 & 0.6385 \\
& \textbf{SurfSurg6D} & \textbf{0.7952} & \textbf{0.6897} \\
\bottomrule
\vspace{-8mm}
\label{tab:2d_projection}
\end{tabular}}
\end{table}

 Our experiments compared SurfSurg6D with    5  representative object pose estimation methods,    three  require training and two claim to be capable of inferring for unseen objects. Among them, GDR-Net\cite{GDR-net} focuses on direct regression based on geometric information, and  MRC-Net applies a first-classify-then-compare paradigm, doing template matching after coarsely regressing the pose.  Surfemb\cite{surfemb} focuses on dense correspondence and surface modeling. MegaPose \cite{megapose} performs zero-shot inference via rendering-and-comparing, and FoundPose \cite{foundpose} aims to extract general patch descriptors from foundation vision models DINOv2 to perform unseen object 2D-3D correspondence.  We also include two leading methods, ICL and IGTUM in the SurgRIPE\cite{SurgRIPE} 
challenge.  The tag \textit{real} indicates that only the SurgRIPE training dataset is used, and the tag \textit{syn+real} indicates that both the SurgRIPE training datasets and generated SynSurg6D datasets are involved in training.

\begin{figure}[!ht]
\centering
\includegraphics[width=0.9\linewidth]{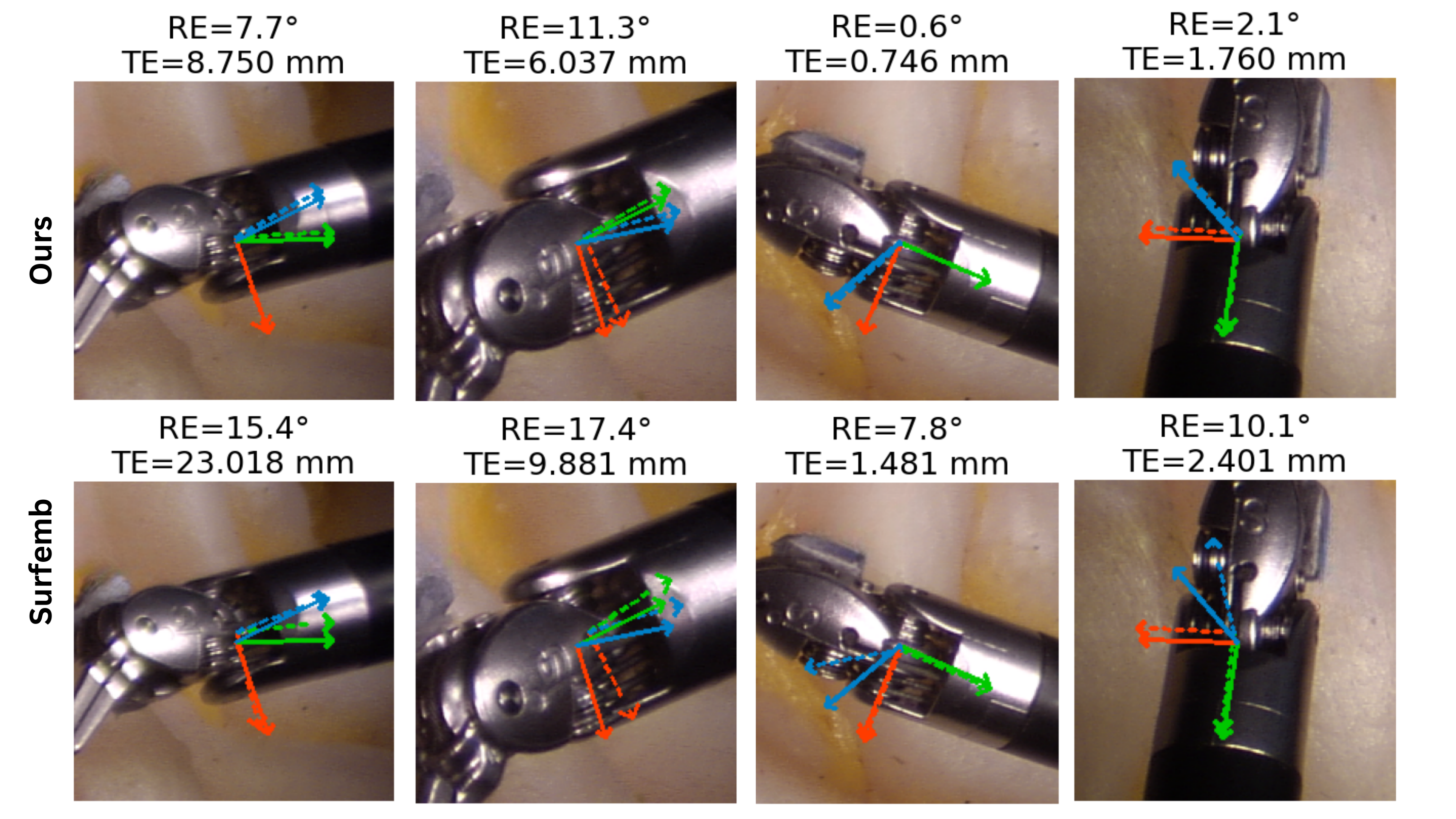}
\caption{Examples of pose estimation results of SurfSurg6D and Surfemb. Solid axis indicates the ground truth, and the dashed axis indicates the prediction. }
\label{fig_pose}
\end{figure}

For 6-DoF pose evaluation with SurgRIPE dataset , as shown in Table \ref{tab:main-results}, we can safely come to the conclusion from the table that the generated SynSurg6D steadily enhances pose estimation approaches' performance, especially in non-occlusion cases  and data-driven methods like MRC-Net. This validates our assumption that in the surgical domain, the scarcity of high-quality annotated data is still the main obstacle for instrument pose estimation. The failure of two models designed for unseen object pose estimation indicates the incapability of foundation models for textureless objects with reflective surfaces, emphasizing the necessity of collecting and generating data of these kinds of objects. The success of SurfSurg6D  (see visualization in Fig.\ref{fig_pose}.)  shows the effectiveness of our hard negative mining strategy for correspondence learning and the regularization for surface modeling.

\begin{figure}[!ht]
\centering
\includegraphics[width=0.9\linewidth]{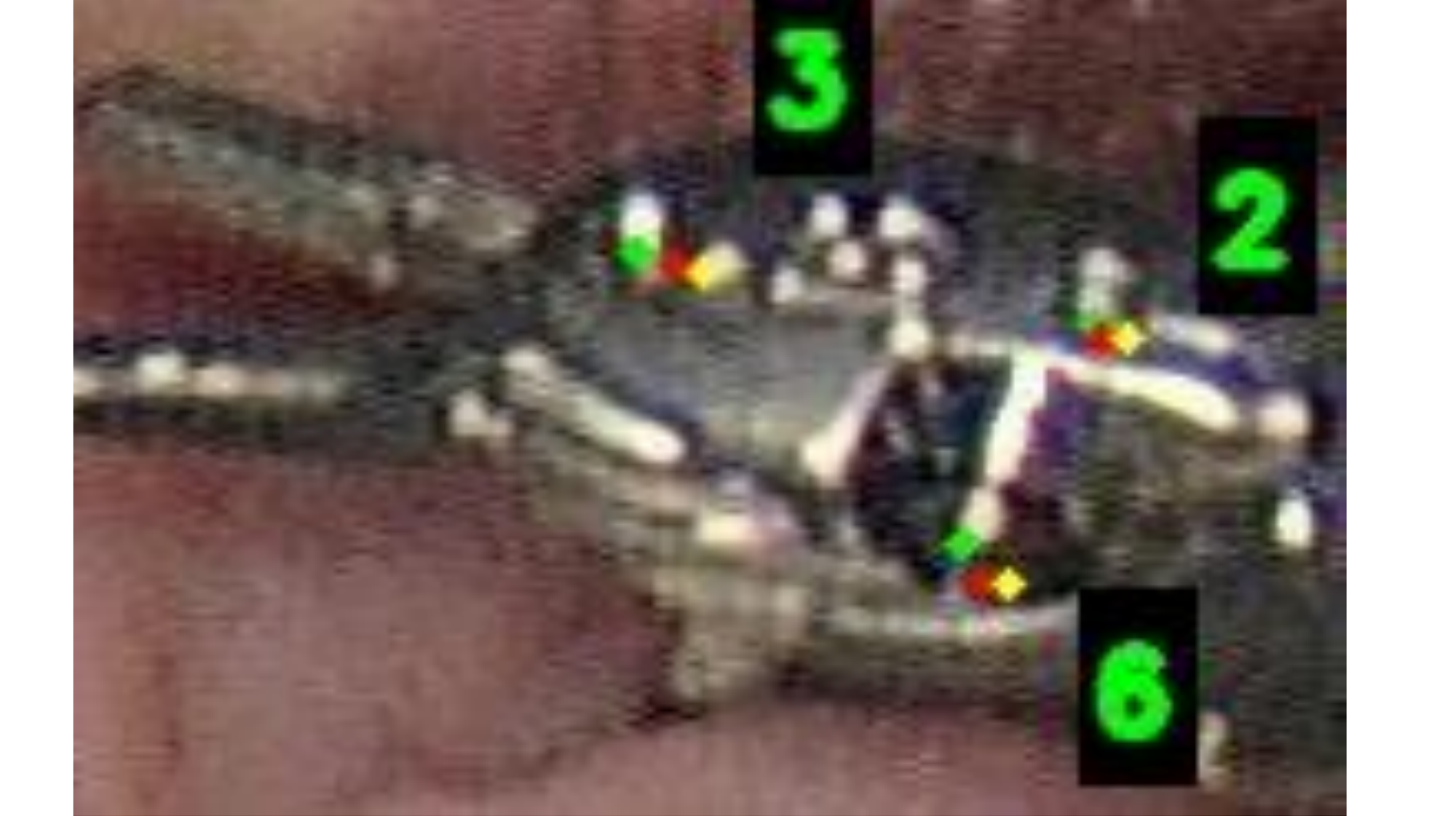}
\caption{The example of keypoint projection results of SurgPose. Green, red and yellow points indicate ground truth, SurfSurg6D and Surfemb.}

\label{fig_surgpose}
\end{figure}

For 2D projection  qualitative evaluation,  we employ SurgPose\cite{surgpose} 
and EndoVis2018\cite{EndoVis2018}, as shown in     Table \ref{tab:2d_projection} and Fig.\ref{fig_surgpose}. The results verify the generalization  of SynSurg6D brings to the model in new scenes.
\vspace{-4mm}

 \subsection{Ablation Study}

In this part, we analyze the effectiveness of both the hard negative pair mining strategy and the consistency loss for surface modeling. The results are shown in Table \ref{tab:abla}.

\begin{table}[!t]
    \centering

    \caption{Ablation experiments about hard negative mining strategy and consistency loss for surface modeling.}  
  \resizebox{\columnwidth}{!}{
        \begin{tabular}{lccc}
        \toprule[1.5pt]
        & \multicolumn{3}{c}{SurgRIPE}  \\ 
        \cmidrule{2-4} 
        Method & RE(\degree) $\downarrow$ &TE(mm)$\downarrow$ & ADD-10 Acc (\%) $\uparrow$  \\ 
        \midrule[1pt]
        Surfemb  & 7.22 & 3.38 & 41.56  \\
        $+$Hard Negative Mining  & 7.03  & 3.20  & 42.06   \\
        $+$ Consistency Loss & 6.71 & 3.30 & 42.38 \\
        \textbf{SurfSurg6D} & \textbf{6.42} & \textbf{3.20} & \textbf{42.89} \\
        \bottomrule[1.5pt]
        \end{tabular}
        }
 
    \label{tab:abla}
\end{table}

From Table~\ref{tab:abla}, hard negative mining mainly improves translation, while consistency regularization benefits rotation. Reweighting near-ambiguous negatives reduces erroneous correspondences around edges and self-occlusions, yielding cleaner RANSAC-PnP inliers and more stable translation. The consistency loss encourages local smoothness in $M_{\phi}$, making surface retrieval more robust to prediction noise and improving orientation accuracy.

Combining both gives the best results by balancing local discrimination and global smoothness. As shown in the subplot in Fig.~\ref{fig:mainfig}, consistency regularization produces a smoother, geometry-aware embedding field, so small embedding deviations still retrieve nearby surface points, leading to more stable 6-DoF pose estimation.

\section{Discussion and Conclusion}
Our work proposes a new synthetic dataset, SynSurg6D, to alleviate the data shortage in the task of surgical instrument pose estimation, as well as two new components that help improve textureless object surface modeling and correspondence learning through hard negative pair mining and embedding consistency regularization. We validate the effectiveness of our contributions by evaluating them on SurgRIPE\cite{SurgRIPE}   ,  EndoVis2018 \cite{EndoVis2018}  and SurgPose\cite{surgpose}
. The results on SurgRIPE show an improvement of 2.45\% in ADD-10 accuracy compared to Surfemb\cite{surfemb} training with the combination of SurgRIPE and SynSurg6D. Results on  EndoVis2018  and SurgPose indicate the method's strong generalization to data in real scenes.

We also evaluate two foundation models that claim to be capable of unseen object pose estimation on SurgRIPE, yet exhibited unsatisfactory performance. These foundation models may lack the ability to tackle textureless objects or occlusion scenes, especially when surfaces are reflective or edge cues are missing. A dataset with variety in these aspects would help models better adapt to these circumstances.

In our data generation process, we find that the synthesized dataset sometimes does not resemble metallic luster or deterioration. 3D Gaussian Splatting    \cite{splattingdataset,instrumentsplatting,diffcompare} 
 tends to tackle these problems better, but the method cannot adapt to randomized lighting conditions. Further exploration will be conducted in this aspect. Our framework currently estimates instrument parts independently.     An important next step is to recover the full articulated configuration (6-DoF base pose + joint angles) of the instrument, which is more informative for the downstream analysis of surgical motions. This would require supervision of multiple keypoints on different parts of instruments, especially those near articulations, as shown in Fig.\ref{fig_surgpose}, yet current method still lacks sufficient precision. More 2D keypoint detection techniques may be introduced to help with this perspective.

 For inference speed, SurfSurg6D and Surfemb achieve 1.25 FPS, FoundPose 0.60 FPS, GDR-Net 1.8 FPS, MRC-Net 6.0 FPS, and MegaPose 0.08 FPS. We currently use the ground truth bounding boxes for all comparisons, so the inference speed might be further slowed if detection modules are introduced. This is one of our current method's limitations, yet it could be improved by changing the inference settings to better adapt to real surgical video streams, where pose changes between adjacent frames are continuous and temporally correlated. A more practical inference pipeline is to use the pose estimator only for initialization and failure recovery, while relying on a lightweight temporal tracking module for fast pose propagation in most frames. Under such a design, the expensive matching and registration steps are invoked only occasionally, which can significantly improve the effective inference speed. Therefore, the final inference speed could be increased to the real-time level even when introduced with bounding box detection modules.

In the future, the work may be extended to instrument pose tracking in surgical videos, which provide crucial cues for high-level tasks, including error detection and instrument-scene interaction. We also plan to analyze the relationship of estimated kinematics data with certain surgical procedures or actions, so that a standardized analysis pipeline for robotic surgical videos can be established.



\bibliographystyle{IEEEtran}
\bibliography{ref}




\vfill

\end{document}